\documentclass[runningheads]{llncs}
\usepackage{graphicx}
\usepackage{tikz}
\usepackage{comment} 
\usepackage{amsmath,amssymb} % define this before the line numbering.
\usepackage{color}
\usepackage[symbol]{footmisc}
\usepackage[font=scriptsize,labelfont=bf]{caption}
\usepackage[T1]{fontenc}

\usepackage[width=122mm,left=12mm,paperwidth=146mm,height=193mm,top=12mm,paperheight=217mm]{geometry}
\usepackage{multirow}
\usepackage[ruled,noend]{algorithm2e}
\usepackage{balance}
\usepackage{enumitem}
\usepackage{times} % this one, solved.
\usepackage{epsfig}
\usepackage{wrapfig}

\usepackage{color}
\definecolor{mygray}{gray}{.6}
\definecolor{dg}{rgb}{0,0.694,0.298}
\definecolor{purple}{rgb}{0.4,0.176,0.569}
\definecolor{pink}{cmyk}{0, 0.7808, 0.4429, 0.1412}
\newcommand{\first}[1]{\textbf{\textcolor{red}{{}#1}}}
\newcommand{\second}[1]{\textbf{\textcolor{green}{{}#1}}}
\newcommand{\third}[1]{\textbf{\textcolor{blue}{{}#1}}}
\usepackage[colorlinks,
            linkcolor=red,
            anchorcolor=blue,
            citecolor=blue
            ]{hyperref}
\newcommand{\felix}[1]{\textbf{\textcolor{red}{Felix: #1}}}

\usepackage{xspace}
\makeatletter
\DeclareRobustCommand\onedot{\futurelet\@let@token\@onedot}
\def\@onedot{\ifx\@let@token.\else.\null\fi\xspace}
\def\eg{\emph{e.g}\onedot} 
\def\ie{\emph{i.e}\onedot}

\makeatother

\begin{document}
% \renewcommand\thelinenumber{\color[rgb]{0.2,0.5,0.8}\normalfont\sffamily\scriptsize\arabic{linenumber}\color[rgb]{0,0,0}}
% \renewcommand\makeLineNumber {\hss\thelinenumber\ \hspace{6mm} \rlap{\hskip\textwidth\ \hspace{6.5mm}\thelinenumber}}
% \linenumbers
\pagestyle{headings}
\mainmatter
\def\ECCVSubNumber{100}  % Insert your submission number here
\setlength{\textfloatsep}{5pt}

\title{SPARK: Spatial-aware Online Incremental Attack Against Visual Tracking}

%===========================================================================%
% INITIAL SUBMISSION 
\begin{comment}
\titlerunning{ECCV-20 submission ID \ECCVSubNumber} 
\authorrunning{ECCV-20 submission ID \ECCVSubNumber} 
\author{Anonymous ECCV submission}
\institute{Paper ID \ECCVSubNumber}
\end{comment}
%******************

% CAMERA READY SUBMISSION
%\begin{comment}
\titlerunning{SPARK: Spatial-aware Online Incremental Attack Against Visual Tracking} 

\author{Qing Guo\inst{1,2\footnotemark[1]} \and
Xiaofei Xie\inst{2\footnotemark[1]}\and
Felix Juefei-Xu\inst{3} \and
Lei Ma\inst{4} \and
Zhongguo Li\inst{1} \and \\
Wanli Xue\inst{5} \and
Wei Feng\inst{1\footnotemark[2]} \and
Yang Liu\inst{2}
}
\authorrunning{Q. Guo et al.}
% First names are abbreviated in the running head.
% If there are more than two authors, 'et al.' is used.
%
\institute{College of Intelligence and Computing, Tianjin University, China \and
Nanyang Technological University, Singapore \and 
Alibaba Group, USA \and
Kyushu University, Japan \and
Tianjin University of Technology, China \\
\email{tsingqguo@gmail.com}}
%\end{comment}
%******************
\maketitle

\renewcommand{\thefootnote}{\fnsymbol{footnote}} 
\footnotetext[1]{Qing Guo and Xiaofei Xie contributed equally to this work.}
\footnotetext[2]{Wei Feng is the corresponding author~(wfeng@tju.edu.cn).}

\begin{abstract}
Adversarial attacks of deep neural networks have been intensively studied on image, audio, natural language, patch, and pixel classification tasks. Nevertheless, as a typical while important real-world application, the adversarial attacks of online video object tracking that traces an object's moving trajectory instead of its category are rarely explored.
In this paper, we identify a new task for the adversarial attack to visual tracking: online generating imperceptible perturbations that mislead trackers along with an incorrect (Untargeted Attack, UA) or specified trajectory~(Targeted Attack, TA).
To this end, we first propose a \textit{spatial-aware} basic attack by adapting existing attack methods, \ie, FGSM, BIM, and C\&W, and comprehensively analyze the attacking performance. 
We identify that online object tracking poses two new challenges: 1) it is difficult to generate imperceptible perturbations that can transfer across frames, and 2) real-time trackers require the attack to satisfy a certain level of efficiency.
To address these challenges, we further propose the \textbf{\underline{sp}atial-\underline{a}ware online inc\underline{r}emental attac\underline{k}} (a.k.a. SPARK) that performs spatial-temporal sparse incremental perturbations online and makes the adversarial attack less perceptible.
In addition, as an optimization-based method, SPARK quickly converges to very small losses within several iterations by considering historical incremental perturbations, making it much more efficient than basic attacks.
The in-depth evaluation of state-of-the-art trackers (\ie, SiamRPN++ with AlexNet, MobileNetv2, and ResNet-50, and SiamDW) on OTB100, VOT2018, UAV123, and LaSOT demonstrates the effectiveness and transferability of SPARK in misleading the trackers under both UA and TA with minor perturbations.
\end{abstract}

%=============================================%
\section{Introduction}\label{sec:intro}

While deep learning achieves tremendous success over the past decade, the recently intensive investigation on image processing tasks \eg, image classification~\cite{Szegedy_2013_arxiv,Goodfellow_2014_arxiv,Dezfooli2016DeepFoolAS}, object detection~\cite{Xie2017AdversarialEF}, and semantic segmentation~\cite{Metzen2017UniversalAP}, reveal that the state-of-the-art deep neural networks (DNNs) are still vulnerable from adversarial examples.
The minor perturbations on an image, although often imperceptible by human beings,
can easily fool a DNN classifier, detector or segmentation analyzer, resulting in incorrect decisions.
This leads to great concerns especially when a DNN is applied in the safety- and security-critical scenarios.
For a particular task, the domain-specific study and the understanding of how adversarial attacks influence a DNN's performance would be a key to reduce such impacts towards further robustness enhancement~\cite{Wei2018SparseAP}.

\begin{figure}[t]
	\begin{center}
		\includegraphics[width=0.98\linewidth]{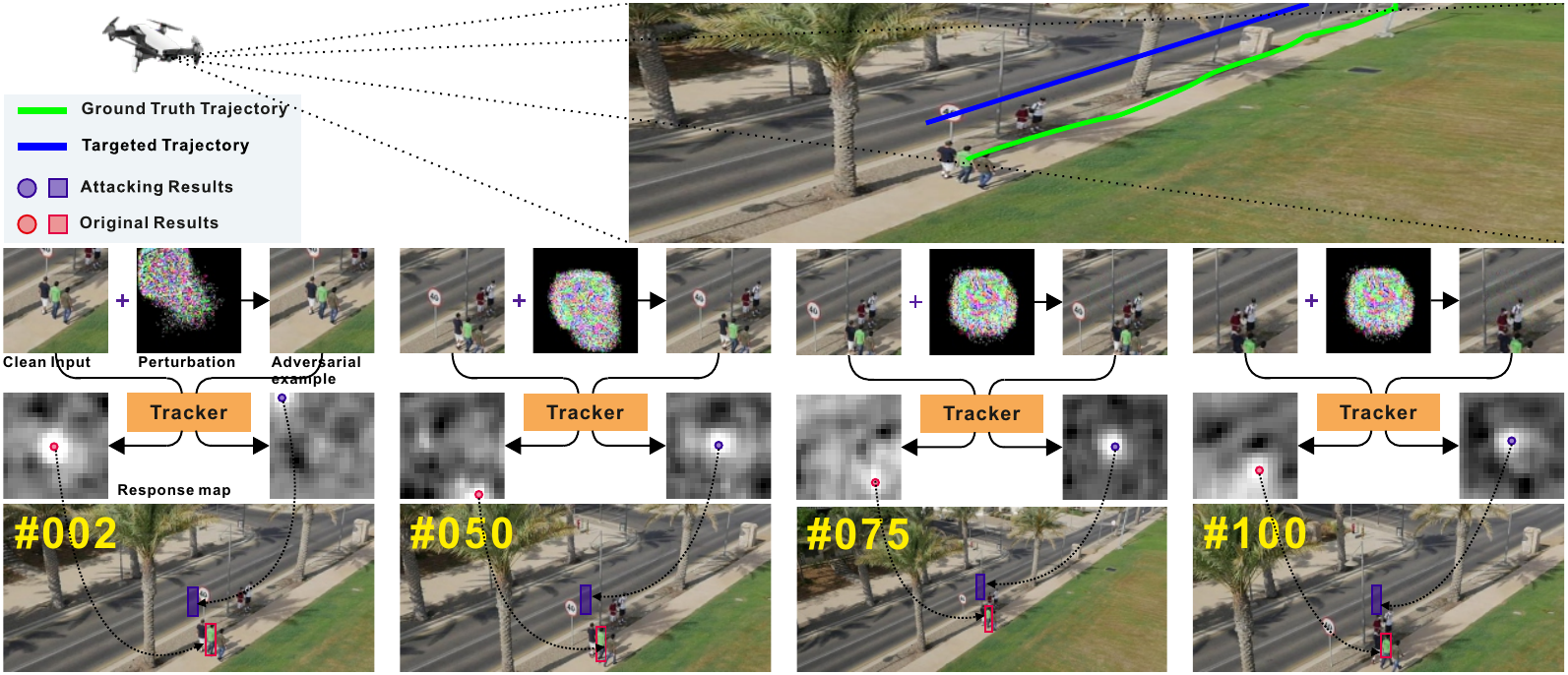}%\vspace{-2em}
	\end{center}
	\caption{\fontfamily{bch}\selectfont {An example of our adversarial attack to online VOT that drives an UAV~\cite{Mueller2018ECCV} to move along the targeted trajectory (the \textcolor{blue}{blue} line), which causes divergence from the object moving path (the \textcolor{green}{green} line). The perturbations are enlarged by $\times 255$ for better visualization.}}
	\label{fig:intro}
	%\vspace{-2em}
\end{figure}

Besides image processing tasks, recent studies also emerge to investigate the adversarial attacks to other diverse types of tasks, \eg, speech recognition~\cite{Carlini2018arxiv,Qin2019arxiv,Cisse2017arxiv}, natural language processing~\cite{Jin2019arXiv,ren-etal-2019-generating,zhang-etal-2019-generating-fluent}, continuous states in reinforcement learning \cite{jianwendrl}, action recognition and object detection~\cite{Wei2018SparseAP,Wei2019TransferableAA}.
Visual object tracking (VOT), which performs online object localization and moving trajectory identification, is a typical while important component in many safety- and security-critical applications, with urgent industrial demands, \eg, autonomous driving, video surveillance, general-purpose cyber-physical systems.
For example, a VOT is often embedded into a self-driving car or unmanned aerial vehicle (UAV) as a key perception component, that drives the system to follow a target object (see Fig.~\ref{fig:intro}).
Adversarial examples could mislead the car or UAV with incorrect perceptions, causing navigation into dangerous environments and even resulting in severe accidents.
Therefore, it is of great importance to perform a comprehensive study of adversarial attacks on visual object tracking.
To this date, however, there exist limited studies on the influence of the adversarial attack on VOT relevant tasks, without which the deployed real-world systems would be exposed to high potential safety risks.

Different from image, speech and natural language processing tasks, online object tracking poses several new challenges to 
the adversarial attack techniques.
% existing attack methods. 
%
\emph{First}, compared with existing sequential-input-relevant tasks, \eg, audios~\cite{Carlini2018arxiv}, natural languages~\cite{Jin2019arXiv} or videos~\cite{Wei2018SparseAP} for classification that have access to the complete sequential data, object tracking processes incoming frames one by one in order.
% while future frame
When a current frame $t$ is under attack, all the previous frames (\ie, $\{1,2\ldots t-1\}$) are already analyzed and cannot be changed. At the same time, the future frames (\ie, $\{t+1,\ldots\}$) are still unavailable and cannot be immediately attacked as well.
% we cannot use future frames or change produced perturbations. 
%
With limited temporal data segments and the dynamic scene changes, it is even more difficult to generate imperceptible yet effective adversarial perturbations that can transfer over time (\ie, multiple consecutive frames).
\emph{In addition}, the object tracking often depends on a target designated object template cropped from the first frame of a video ~\cite{Bertinetto16-2,Li2018CVPR} for further analysis. The different initially designated object might lead to different tracking analysis, which renders the universal adversarial perturbation \cite{Dezfooli2016DeepFoolAS} often ineffective.

\emph{Furthermore}, object tracking usually functions at real-time speed. Thus, it requires the attacks to be efficient enough so that the adversarial perturbation of the current frame can be completed before the next frame arrives. 
Although the gradient descent-based methods (\eg, FGSM~\cite{Goodfellow_2014_arxiv}, BIM~\cite{BIM_2016_ICLRW}) are demonstrated to be effective in attacking the image classifier, they still encounter efficiency issues in fooling the state-of-the-art trackers when multiple frames quickly arrive.

It is also rather expensive for attacking on multiple frames in real-time~\cite{Wei2018SparseAP}. 

To better understand the challenges and uniqueness in attacking the VOT, we first propose a \textit{spatial-aware} basic attack method by adapting the existing state-of-the-art attacking techniques (\ie, FGSM, BIM, C\&W) that are used to attack each frame individually.
Our empirical study confirms that the basic attack is indeed ineffective for attacking the VOT, due to the consecutive temporal frames in real-time.
Based on this, we further propose the \textit{\underline{sp}atial-\underline{a}ware online inc\underline{r}emental attac\underline{k}} (SPARK) method that can generate more imperceptible perturbations online in terms of both effectiveness and efficiency.
%
%------------------------------------
%------------------------------------
%
The main contributions of this paper are as follows:%\vspace{-0.7em}
%------------------------------------
\begin{itemize}
	\setlength{\itemsep}{0pt}
    \setlength{\parsep}{0pt}
    \setlength{\parskip}{0pt}    
	\item We formalize the adversarial attack problem for the VOT, \ie, 
	generating imperceptible perturbations online to mislead visual trackers that traces an object, into an incorrect (Untargeted Attack, UA) or specified (Targeted Attack, TA) trajectory.
	
	\item We propose several \textit{basic attacks} by adapting existing attacks (\ie, FGSM, BIM, C\&W) and further perform an empirical study for better understanding challenges of adversarial attacks on real-time object tracking.
	
	\item We propose a new \textit{\underline{sp}atial-\underline{a}ware online inc\underline{r}emental attac\underline{k}} (SPARK) method that can efficiently generate imperceptible perturbations for real-time VOT.
	
	\item In line with the basic methods, our in-depth evaluation demonstrates the effectiveness and efficiency of SPARK in attacking the state-of-the-art SiamRPN++ trackers with AlexNet, MobileNetv2, and ResNet-50 models \cite{Li2018CVPR,Li2019CVPR} and SiamDW trackers \cite{Zhang2019CVPR} under UA and TA. The generated attacks also exhibit strong transferability to the online updating variants of SiamRPN trackers. 
	
\end{itemize}
%------------------------------------

%-------------------------------------------------------------------------
%-------------------------------------------------------------------------
%\section{Related Work}
%\vspace{-1.5em}
\section{Related Work}
%\vspace{-0.5em}
%
%-------------------------------------------------------------------------
%\subsection{Adversarial Examples}
\textbf{Adversarial Examples.}
Extensive studies have shown the vulnerability of DNN from adversarial attacks \cite{Ling2019DEEPSECAU}.
\cite{Szegedy_2013_arxiv} initially shown the existence
of adversarial attacks, and \cite{Goodfellow_2014_arxiv} proposed an efficient one-step method FGSM, that was later improved via iterative method~\cite{BIM_2016_ICLRW} and momentum term~\cite{Dong2018BoostingAA}. 
Similarly, \cite{Papernot2016TheLO} proposed the Jacobian-based saliency map attack with high success rate, while \cite{CW_2017_SSP} realized effective attack by optimization methods (C\&W) under different norms. 
Further adversarial attacks were extended to tasks like object detection \cite{Xie2017AdversarialEF,Li2018RobustAP,Zhao2018Arxiv}, semantic segmentation \cite{Xie2017AdversarialEF,MoosaviDezfooli2017UniversalAP},  and testing techniques for DNNs \cite{ma2018deepgauge,deephunter19,deepstellar}.

Recent works also confirmed the existence of adversarial examples in sequential data processing, \eg, speech recognition \cite{Cisse2017arxiv,Carlini2018arxiv,Qin2019arxiv}, natural language \cite{Gao2018SPW,Jin2019arXiv}, and video processing \cite{Wei2018SparseAP}. 
Different from these works, our attack aims at misleading trackers with limited online data access, \ie, the future frames are unavailable, the past frames cannot be attacked either.
Among the most relevant work to ours, 
\cite{Wei2018SparseAP} proposed the $L_{2,1}$ norm-based attack to generate sparse perturbations for action recognition, under the condition that the whole video data is available and the perturbations of multiple frames can be jointly tuned.
To further show the difference, we implement a tracking attack with \cite{Wei2018SparseAP} and compare it in the evaluation.
\cite{Li2018RobustAP} attacked the region proposal network (RPN) that is also used in the SiamRPN trackers~\cite{Li2018CVPR}. However, this attack focuses on fooling image detectors to predict inaccurate bounding boxes, thus cannot be directly used to attack trackers aiming to mislead to an incorrect trajectory with online videos.
\cite{Wei2019TransferableAA} proposed the video object detection attack by addressing each frame independently, which is not suitable for online tracking where the tracker often runs at real-time speed.
Another related work \cite{Lin2017IJCAI} studied when to attack an agent in the reinforcement learning context. In contrast, this work mainly explores how to use temporal constraints to online generate imperceptible and effective perturbations to mislead real-time trackers.
%\textcolor{red}{In addition, some testing techniques~\cite{ma2018deepgauge,deephunter19,deepstellar} have been proposed to generate adversarial examples for testing DNNs with the guidance of coverage criteria.} 

%-------------------------------------------------------------------------
%\subsection{Visual Object Tracking}
\textbf{Visual Object Tracking}
Visual tracking is a fundamental problem in computer vision, estimating positions of an object (specified at the first frame) over frames \cite{Wu15}.
The state-of-the-art trackers can be roughly summarized to three categories, including correlation filter-based \cite{Danelljan16ECO,AL17-csrdcf,ChenICME2018,Zhang_neuro2019,DSARCF_TIP2019,Guo_TIP2020}, classification \& updating-based \cite{MDNet16,SCT_TIP2017,Song2018CVPR} and Siamese network-based trackers \cite{Bertinetto16-2,Guo17_ICCV,Zhu2018ECCV,Wang2018CVPR,Wang2019CVPR,Fan2019CVPR}.
Among these works, Siamese network-based methods learn the matching models offline and track objects without updating parameters, which well balances the efficiency and accuracy. 
In particular, the SiamRPN tracker can adapt objects' aspect ratio changing and run beyond real time~\cite{Li2018CVPR}.
In this paper, we choose SiamRPN++ \cite{Li2019CVPR} with AlexNet, MobileNetv2, and ResNet-50 as subject models due to following reasons: 1) SiamRPN++ trackers are widely adopted with high potential to real-world applications~\cite{Kristan2018a,Li2019CVPR}. 
The study of attacking to improve their robustness is crucial for industrial deployment with safety concerns.
2) Compared with other frameworks (\eg, correlation filter-based trackers), SiamRPN is a near end-to-end deep architecture with fewer hyper-parameters, making it more suitable to investigate the attacks. In addition to SiamRPN++, we attack another state-of-the-art tracker, \ie, SiamDW \cite{Zhang2019CVPR}, to show the generalization of our method.

\begin{figure}[t]
	\begin{center}
		\includegraphics[width=0.98\linewidth]{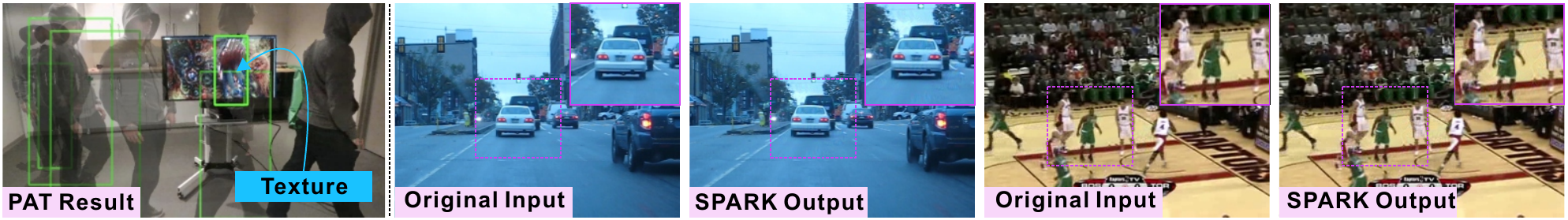}%\vspace{-2em}
	\end{center}
	\caption{\fontfamily{bch}\selectfont {Difference between PAT~\cite{Wiyatno2019Arxiv} and our method. PAT produces \textit{perceptible pattern}.and let the GOTURN tracker lock on the texture. The adversarial perturbations of SPARK are imperceptible and hardly perceived.}}
	\label{fig:diff}
	%\vspace{-2em}
\end{figure}

%-------------------------------------------------------------------------
%\subsection{Difference to PAT~\cite{Wiyatno2019Arxiv}}
\textbf{Difference to PAT~\cite{Wiyatno2019Arxiv}.}
To the best of our knowledge, until now, there has been a limited study on attacking online object tracking. 
\cite{Wiyatno2019Arxiv} generated physical adversarial textures~(PAT) via white-box attack to let the GOTURN tracker~\cite{Held2016ECCV} lock on the texture when a tracked object moves in front of it. 
The main differences between our method and PAT are:
{\bf (1)} Their attack objectives are distinctly and totally different. As shown in Fig.~\ref{fig:diff}, PAT is to generate \textit{perceptible texture} and let the GOTURN tracker lock on it while our method is to online produce \textit{imperceptible perturbations} that mislead state-of-the-art trackers, \eg, SiamRPN++~\cite{Li2019CVPR}, along an incorrect or specified trajectory. 
{\bf (2)} Different theoretical novelties. PAT is to improve an existing Expectation Over Transformation~(EOT)-based attack by studying the need to randomize over different transformation variables. Our work intends to perform a comprehensive study on adapt existing adversarial attacks on object tracking and reveal the new challenges in this important task. We then proposed a novel method, \ie, spatial-aware online incremental attack, which can address these challenges properly.
{\bf (3)} Different subject models. PAT validates its method by attacking a light deep regression tracker, \ie, GOTURN that has low tracking accuracy on modern benchmarks~\cite{Fan2019LaSOT,Wu15,Kristan2018a}. We use our method to attack the state-of-the-art trackers, \eg, SiamRPN++~\cite{Li2019CVPR} and SiamDW~\cite{Zhang2019CVPR}.

%------------------------------------
%------------------------------------
%-------------------------------------------------------------------------
%-------------------------------------------------------------------------
%\vspace{-1em}
\section{Spatial-aware Online Adversarial Attack}
%
%In this section, we first extend adversarial examples to single object tracking and aim to online generate imperceptible perturbations that can mislead trackers along a incorrect or specified trajectory.
%
%To this end, an online incremental attack method is introduced for online video perturbations.

%-------------------------------------------------------------------------
%\vspace{-1em}
\subsection{Problem Definition}
%\vspace{-0.5em}
%
Let $\mathcal{V}=\{\mathbf{X}_t\}_1^T$ be an online video with $T$ frames, where $\mathbf{X}_t$ is the $t$th frame. 
Given a tracker $\phi_\theta(\cdot)$ with parameters $\theta$, we crop an object template $\mathbf{T}$ (\ie, the target object) from the first frame. 
The tracker is tasked to predict bounding boxes that tightly wrap the object in further incoming frames.

To locate the object at frame $t$, the tracker calculates $\{(y^i_t,\mathbf{b}^i_t)\}_{i=1}^N=\phi_\theta(\mathbf{X}_t,\mathbf{T})$,
where $\{\mathbf{b}^i_t\in\Re^{4\times 1}\}_{i=1}^N$ are $N$ object candidates in $\mathbf{X}_t$ and $y^i_t$ indicates the positive activation of the $i$th candidate (\ie, $\mathbf{b}^i_t$).
We denote the tracker's predictive bounding box of the target object at the clean $t$th frame by $\mathbf{b}_t^\mathrm{gt}\in\Re^{4\times 1}$ and the object tracker assigns the predictive result $OT(\mathbf{X}_t,\mathbf{T})= {\mathbf{b}^\mathrm{gt}_t} = {\mathbf{b}^k_t}$, where $k=\arg\max_{1\le i\le N}(y^i_t)$, \ie, the bounding box with highest activate value is selected as the \textit{predictive object} at frame $t$.
The above tracking process covers most of the state-of-the-art trackers, \eg, Siamese network-based trackers~\cite{Zhang2018ECCV,Dong2018ECCV,Fan2019CVPR,Zhang2019CVPR,Li2019CVPR,He2018CVPR,Bertinetto16-2} and correlation filter-based trackers~\cite{Dai2019CVPR,Sun2019CVPR,Guo17_ICCV}.
We define the adversarial attacks on tracking as follows:

{\bf Untargeted Attack~(UA).} 
	UA is to generate adversarial examples $\{\mathbf{X}_t^\mathrm{a}\}_1^T$ such that 
	% 	$\{y^\mathrm{a}_t\neq y^\mathrm{gt}_t\}_1^{T}$ and 
	$\forall1\le t\le T$, $\mathrm{IoU}(OT(\mathbf{X}_t^\mathrm{a},\mathbf{T}), \mathbf{b}^\mathrm{gt}_t)=0$, where  %$(y^\mathrm{a}_t,\mathbf{b}^\mathrm{a}_t)=\arg\max_{y^i_t}\phi_\theta(\mathbf{X}_t^\mathrm{a},\mathbf{T})$ 
	$\mathrm{IoU}(\cdot)$ is the Intersection over Union between two bounding boxes. 

{\bf Targeted Attack~(TA).} 
	Suppose a \textit{targeted trajectory} $\{\mathbf{p}^\mathrm{tr}_t\}_1^T$ desires the trajectory we hope the attacked tracker to output, \eg, the blue line in Fig.~\ref{fig:intro}.
	%\felix{'desire the targeted tracking position', not so clear to me. Switch an expression.} the targeted tracking position at each frame.
	TA is to generate adversarial examples $\{\mathbf{X}_t^\mathrm{a}\}_1^{T}$ such that $\forall1\le t\le T$, $ce(OT(\mathbf{X}_t^\mathrm{a},\mathbf{T})) =\mathbf{p}^\mathrm{tr}_t$, where $ce(\cdot)$ shows the center position of the bounding box and $\mathbf{p}^\mathrm{tr}_t$ depicts the targeted position at the $t$th frame.

%\fei{It is better to have some intuitive explanation for the two goals.}
Intuitively, UA is to make the trackers predict incorrect bounding boxes of a target object at all frames by adding small distortions to online captured frames while TA aims to intentionally drive trackers to output desired object positions specified by the \textit{targeted trajectory}.

\begin{figure*}[t]
	\begin{center}
		\includegraphics[width=0.98\linewidth]{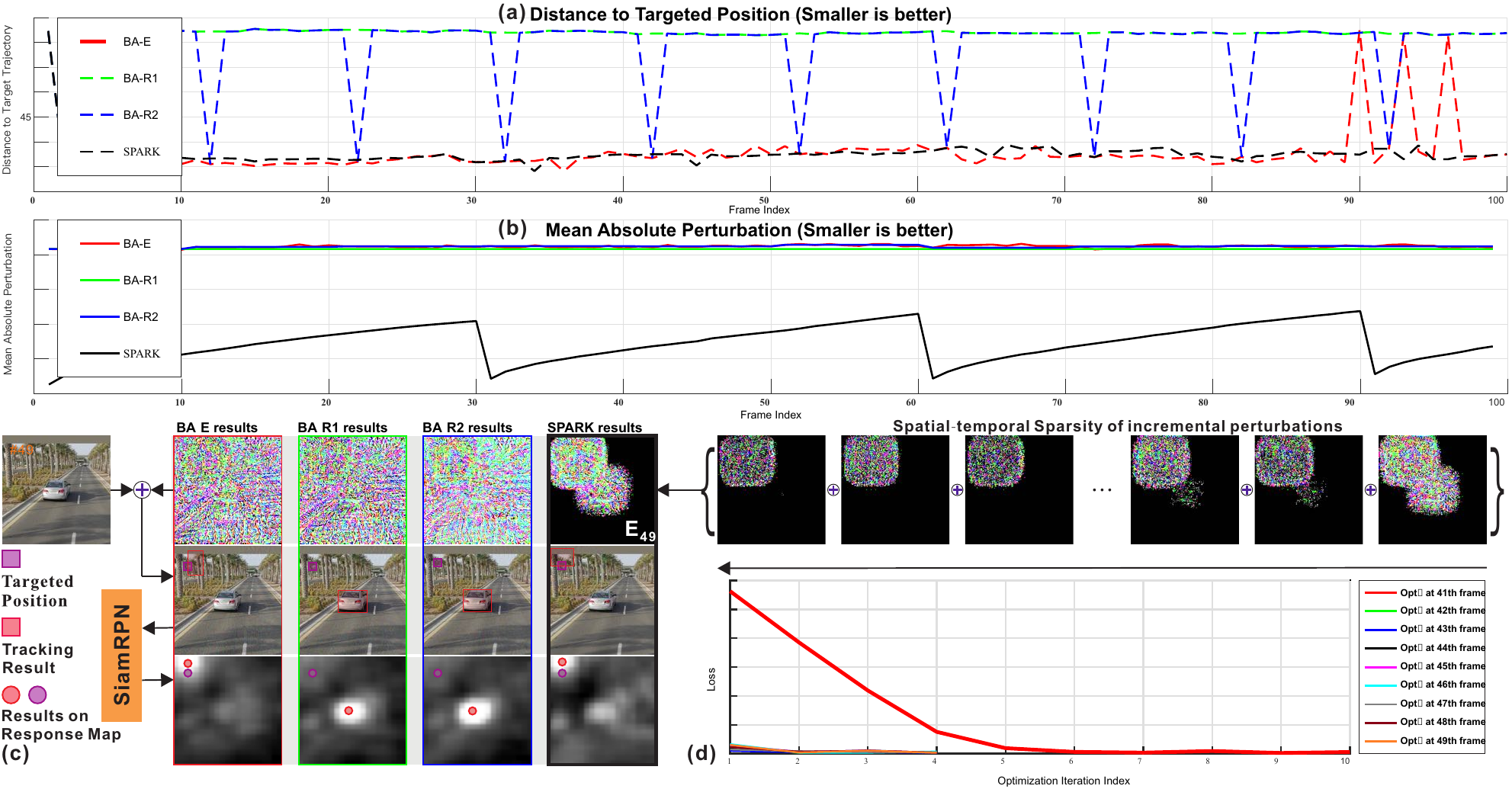}%\vspace{-2em}
	\end{center}
	\caption{\fontfamily{bch}\selectfont{Analysis of our basic attack~(BA) and spatial-aware online incremental attack~(SPARK). (a) shows the distance between the targeted position and predicted object position after attacking. A smaller distance means the attack is more effective. (b) shows the mean absolute perturbation of each frame. A smaller MAP leads to less imperceptible perturbation. (c) presents the adversarial perturbations of 4 attack methods at frame 49, corresponding adversarial examples, and response maps from SiamRPN-AlexNet. (d) includes the incremental perturbations from frame 41 to 49 and the loss values at each frame. The perturbations are enlarged by $\times 255$ for better visualization.}}
	\label{fig:analysis}%\vspace{-2em}
\end{figure*}

%\vspace{-1em}
\subsection{Basic Attack}
%\vspace{-0.5em}
% Based on the definition of tracking attack, one intuitive way is to adopt the state-of-the-art adversarial attacks (e.g., FGSM~\cite{Goodfellow_2014_arxiv}, BIM~\cite{BIM_2016_ICLRW} and C\&W~\cite{CW_2017_SSP}) on the object trackers. Hence, 
We first propose the basic attacks by adapting existing adversarial methods at each frame. To attack a tracker $OT(\cdot)$, we can use \if 0 \felix{adapt, right?} \fi another tracker $OT{'}(\cdot)$ to generate adversarial examples.
{	% \color{red}
For untargeted attack~(UA), at frame $t$, we formally define the problem of finding an adversarial example as follows:
	\begin{align}
	\text{minimize } \;& \mathcal{D}(\mathbf{X}_t, \mathbf{X}_t+\mathbf{E}_t)\\
	\text{subject to~} \;& \mathrm{IoU}(OT{'}(\mathbf{X}_t+\mathbf{E}_t,\mathbf{T}), \mathbf{b}_t^\mathrm{gt{'}})=0 
	\end{align}
	where $\mathbf{X}_t^{\mathrm{a}}=\mathbf{X}_t+\mathbf{E}_t$ and $\mathbf{E}_t$ is the desired distortion that changes the result of the tracker and $\mathcal{D}$ is a distance metric. 
	We follow the setup of FGSM and use the $L_\infty$ norm as $\mathcal{D}$.
	We use $\mathbf{b}_t^\mathrm{gt{'}}$ as the predictive result on the clean frame $\mathbf{X}_t$.
	When $OT(\cdot)=OT{'}(\cdot)$, we consider the attack as a white-box attack.
	
	To achieve the UA, we define the objective function $f^{\mathrm{ua}}$ such that  $ \mathrm{IoU}(OT{'}(\mathbf{X}_t+\mathbf{E}_t,\mathbf{T}),\mathbf{b}^\mathrm{gt{'}})=0$ if and only if $f^{\mathrm{ua}}(\mathbf{X}_t+\mathbf{E}_t,\mathbf{T})<0$:
	\begin{align}
	\label{eq:ua_label}
	f^{\mathrm{ua}}(\mathbf{X}_t+\mathbf{E}_t,\mathbf{T}) = y^{\mathrm{gt}{'}}_t-\max_{\mathrm{IoU}(\mathrm{b}^i_t, \mathrm{b}^{gt{'}}_t)=0} (y^i_t)
	\end{align}
	where $\{(y^i_t,\mathbf{b}^i_t)\}_{i=1}^N=\phi_{\theta{'}}(\mathbf{X}_t+\mathbf{E}_t,\mathbf{T})$, $\theta{'}$ denotes parameters of $OT{'}(\cdot)$, and $y^{\mathrm{gt}{'}}_t$ is the activation value of $\mathbf{b}^{gt{'}}_t$ at the perturbed frame $t$.
	For the targeted attack~(TA), at frame $t$, we define the problem of finding a targeted adversarial example as follows:
	\begin{align}
	\text{minimize } \;& \mathcal{D}(\mathbf{X}_t, \mathbf{X}_t+\mathbf{E}_t)\\
	\text{subject to~} \;& ce(OT{'}(\mathbf{X}_t+\mathbf{E}_t,\mathbf{T})) =  \mathbf{p}^\mathrm{tr}_t
	\end{align}
	where $\mathbf{p}^\mathrm{tr}_t$ is the targeted position at frame $t$ and $ce(\cdot)$ outputs the center position of a bounding box. To achieve the goal, we define the objective function $f^{\mathrm{ta}}$ such that $ce(OT{'}(\mathbf{X}_t+\mathbf{E}_t,\mathbf{T})) =  \mathbf{p}^\mathrm{tr}_t$ if and only if $f^{\mathrm{ta}}(\mathbf{X}_t+\mathbf{E}_t,\mathbf{T})<0$:
	\begin{align}
	\label{eq:ta_label}
	f^{\mathrm{ta}}(\mathbf{X}_t+\mathbf{E}_t,\mathbf{T}) = y^{\mathrm{gt}{'}}_t-\max_{ce(\mathbf{b}^i_t)=\mathbf{p}^\mathrm{tr}_t} (y^i_t)
	\end{align}

	To perform the basic tracking attack, FGSM~\cite{Goodfellow_2014_arxiv}, BIM~\cite{BIM_2016_ICLRW} and C\&W~\cite{CW_2017_SSP} are adapted to optimize the objective functions (\ie, Eq.~(\ref{eq:ua_label}) and Eq.~(\ref{eq:ta_label})).
	% For example, with FGSM, we generate the perturbation by:
	In this paper, we mainly focus on the white-box attack on visual object tracking by setting $OT(\cdot)=OT{'}(\cdot)$ while studying the transferability of different trackers in the experiments.
}
% select the targeted position $\mathbf{p}^\mathrm{tr}_t$ and use the following loss function to generate TA perturbations.
% For each frame, we use the cross-entropy loss function $L(\{\hat{y}_t^i\}_{i=1}^N,\phi_\theta(\mathbf{X}_t+\mathbf{E}_t,\mathbf{T}))$ and generate UA perturbations by solving
% 

% \vspace{0em}\begin{equation}\label{eq:ua_label}
% \hat{y}_t^i=
% \left\{
% 	\begin{aligned}
% 	1 &  &\mathrm{rand}(\{\mathbf{b}_t^i|\mathrm{IoU}(\mathbf{b}^i_t, \mathbf{b}^\mathrm{gt}_t)=0\})\\
% 	0 &  &\mathrm{otherwise}.
% 	\end{aligned}
% \right.
% \end{equation}

% $L(\{\hat{y}_t^i\}_{i=1}^N,\phi_\theta(\mathbf{X}_t+\mathbf{E}_t,\mathbf{T}))$
% \vspace{0em}\begin{equation}\label{eq:ba_loss}
% \mathbf{E_t} = \arg\min_{\mathbf{E}} L(\{\hat{y}_t^i\}_{i=1}^N,\phi_\theta(\mathbf{X}_t+\mathbf{E},\mathbf{T})).
% \end{equation}
% %

% %
% \vspace{0em}\begin{equation}\label{eq:ta_label}
% \hat{y}_t^i=
% \left\{
% \begin{aligned}
% 1 &  &\mathbf{p}_t^i=\mathbf{p}_t^\mathrm{tr},\\
% 0 &  &\mathrm{otherwise}.
% \end{aligned}
% \right.
% \end{equation}\vspace{0em}
% %
% where $\mathbf{p}_t^i$ is the center position of $i$th candidate $\mathbf{b}_t^i$.

%-------------------------------------------------------------------------
%\vspace{-1em}
\subsection{Empirical Study}
%\vspace{-0.5em}
{
	% \color{red}
	In the following, we perform an empirical study on evaluating the effectiveness of the basic attack. In particular, we study two research questions: 1) how effective is the attack by applying basic attack on each frame? 2) how is its impact of the temporal frames in the video?
	To answer the questions, we perform two kinds of basic targeted attacks on a state-of-the-art tracker, \ie, SiamRPN-AlexNet\footnote{We select SiamRPN-AlexNet, since it is a representative Siamese network based-tracker and achieves high accuracy on modern tracking benchmarks with beyond real-time speed.}:
	
     {\bf BA-E:} Online attacking each frame by using FGSM, BIM, and C\&W to optimize Eq.~(\ref{eq:ta_label}), respectively. 
		
	{\bf BA-R:} Randomly select some frames and perform the basic attack on these frames using FGSM, BIM, and C\&W. For frames between two selected frames, we use the perturbation from the first selected one to distort frames in the interval and see if basic attacks could transfer across time. For example, we attack $1$st and $10$th frames with basic attacks while distorting the $2$th to $9$th frames with the perturbation of $1$st frame.
	
	Note that BA-E and BA-R can answer the two questions, respectively. To be specific, we have configured two BA-R attacks. 
	First, each frame is selected to be attacked with a probability $0.1$ (denoted as \textbf{BA-R1}). Second, we perform the basic attack with an interval $10$, \ie, attack at the $1$th, $11$th, $21$th, $\ldots$ frame (denoted as \textbf{BA-R2}).
}

\begin{table*}[t]
    \fontfamily{bch}\selectfont
	\caption{\fontfamily{bch}\selectfont {Comparing basic attacks, \ie, BA-E, BA-R1, and BA-R2 with our SPARK under TA on the OTB100 dataset.}}%\vspace{-1.0em}
	\tiny
	\setlength{\tabcolsep}{1.5pt}
	\begin{center}
		\begin{tabular}{l|c|c|c|c|c|c|c|c|c|c} 
			\hline
			\multirow{2}{*}{} & \multicolumn{3}{|c|}{BA-E} & \multicolumn{3}{|c|}{BA-R1} & \multicolumn{3}{|c|}{BA-R2} & \multirow{2}{*}{SPARK}\\
			\cline{2-10}
			& FGSM & BIM & C\&W & FGSM & BIM & C\&W & FGSM & BIM & C\&W & \\
			\hline
			\hline
			Succ. Rate (\%)& 8.0 & 69.6 & 57.7 & 6.6 & 17.8 & 17.5  & 6.7 & 53.7  & 23.5 & 78.9  \\
			\hline
			Mean Absolute Perturbation & 1.24 & 5.88 & 1.31 & 1.23 & 5.96 & 0.26 & 1.23 & 3.36 & 1.27 & 1.04 \\
			\hline
			Aver. Iter. Num per frame & 1 & 10 & 10 & 0.10 & 0.95 & 0.94 & 0.10 & 4.6 & 4.6 & 2.25 \\
			\hline
			Aver. Cost per frame~(ms) & 56.2 & 326.0 & 264.0 & 5.50 & 39.1 & 24.8 & 5.68 & 189.5 & 121.4 & 62.1 \\
			\hline
		\end{tabular}
	\end{center}
	\label{tab:analysis_results}%\vspace{-6.0em}
\end{table*}
	
	Table~\ref{tab:analysis_results} shows the success rate, mean absolute perturbation, and average iteration per frame of BA-E, BA-R1, and BA-R2 for attacking SiamRPN-AlexNet-based tracker on OTB100 under TA.
	We see that: 1) BA-E methods via BIM and C\&W get high success rate by attacking each frame. Nevertheless, their perturbations are large and attacking each frame with 10 iterations is time-consuming and beyond real-time tracker. Although FGSM is efficient, its success rate is much lower. 
	2) Randomly attacking 10\% frames, \ie, BA-R1, is about 10 times faster than BA-E. However, the success rate drops significantly. 
	3) BA-R2 method attacking at every 10 frames is efficient while sacrificing the success rate. 
	Compared with BA-R1, with the same attacking rate, \ie, 10\% frames, BA-R2 has higher success rate 
	than BA-R1. 
	For example, base on BIM, BA-R2 has over two times larger success rate. It infers that perturbations of neighbor 10 frames have some transferability due to the temporal smoothness.

	A case study based on BIM is shown in Fig.~\ref{fig:analysis}, 
	where we use the three BA attacks to mislead the SiamRPN-AlexNet to locate an interested object at the top left of the scene~(targeted position in Fig.~\ref{fig:analysis}~(c)).
	Instead of following the standard tracking pipeline, we crop the frame according to the ground truth and get a region where the object are always at the center.
	We show the distance between the targeted position (Fig.~\ref{fig:analysis}~(a)) and tracking results, and the mean absolute perturbation (MAP) (Fig.~\ref{fig:analysis}~(b)) at frame level.
	We reach consistent conclusion with Table~\ref{tab:analysis_results}. As the simplest solution, BA-E attacks the tracker successfully at some time~(distance to the targeted position is less than 20) with the MAP around 5.
	However, the attack is inefficient and not suitable for real-time tracking.
	In addition, according to Fig.~\ref{fig:analysis}~(c), the perturbations are large and perceptible. The results answer the first question: attacking on each frame is not effective, \ie, time-consuming and bigger MAP.
	
	Consider the temporal property among frames, if the attack can be transferred between the adjacent frames, we could only attack some frames while reducing the overhead, \eg, BA-R1 and BA-R2. 
	Unfortunately, the results in Table~\ref{tab:analysis_results} and Fig.~\ref{fig:analysis} show that BA-R1 and BA-R2 only work at the specific frames on which the attacks are performed.
	
	The results answer the second question: the perturbations generated by BA is difficult to transfer to the next frames directly due to the dynamic scene in the video (see the results from BA-R1 and BA-R2).

%\vspace{-1.5em}
\subsection{Online Incremental Attack}
%\vspace{-0.5em}
	\label{subsec:oia}
	Base on the empirical study results from basic attacks, we identify that attacking on each frame directly is not effective. As the frames are sequential and the nearby frames are very similar, our deep analysis found that transferability exists between nearby frames.
	However, how to effectively use the perturbations from previous frames while being imperceptible when we attack a new coming frame is questionable.
	A straightforward way is to add previous perturbations to a new calculated one, which will increase the success rate of attacking but lead to significant distortions.
	To solve this problem, we propose \textit{spatial-aware online incremental attack~(SPARK)} that generates more imperceptible adversarial examples more efficiently for tracking. 
	The intuition of SPARK is that we still attack each frame, but apply previous perturbations on the new frame combined with small but effective \textit{incremental perturbation} via optimization.
	
	At frame $t$, the UA with SPARK is formally defined as:
	\begin{align}
	\text{minimize } \;& \mathcal{D}(\mathbf{X}_t, \mathbf{X}_t+\mathbf{E}_{t-1}+\epsilon_t)\\
	\text{subject to~} \;& \mathrm{IoU}(OT{'}(\mathbf{X}_t+\mathbf{E}_{t-1}+\epsilon_t,\mathbf{T}),\mathbf{b}_t^\mathrm{gt{'}})=0 
	\end{align}
	where $\mathbf{E}_{t-1}$ is the perturbation of the previous frame (\ie, $t-1$th fame) and $\epsilon_t$ is the incremental perturbation. Here, the `incremental' means $\epsilon_t=\mathbf{E}_t-\mathbf{E}_{t-1}$, and we further have $\mathbf{E}_t = \epsilon_t+\sum_{t_0}^{t-1}\epsilon_\tau$, where $t_0=t-L$ and $\{\epsilon_\tau\}_{t-L}^{t-1}$ are $L-1$ previous incremental perturbations, and $\epsilon_{t_0}=E_{t_0}$. We denote $t_0=t-L$ as the start of an attack along the timeline. %
	Based on Eq.~\ref{eq:ua_label}, we introduce a new objective function by using $L_{2,1}$ norm to regularize $\{\epsilon_\tau\}_{t_0}^t$ that leads to small and spatial-temporal sparse $\epsilon_t$.
	\begin{align}
	\label{eq:uaoia_label}
	f^{\mathrm{ua}}(\mathbf{X}_t+\epsilon_t+\sum_{t-L}^{t-1}\epsilon_\tau,\mathbf{T})+\lambda\|\Gamma\|_{2,1},
	\end{align}
	where $\Gamma=[\epsilon_{t-L},...,\epsilon_{t-1},\epsilon_t]$ is a matrix that concatenates all incremental values.
	
	Similarly, the TA with SPARK is formally defined as:
	\begin{align}
	\text{minimize } \;& \mathcal{D}(\mathbf{X}_t, \mathbf{X}_t+\mathbf{E}_{t-1}+\epsilon_t)\\
	\text{subject to~} \;& ce(OT{'}(\mathbf{X}_t+\mathbf{E}_{t-1}+\epsilon_t,\mathbf{T})) =  \mathbf{p}^\mathrm{tr}_t.
	\end{align}
	We also modify the objective function Eq.~\ref{eq:ta_label} by adding the $L_{2,1}$ norm and obtain
	\begin{align}
	\label{eq:taoia_label}
	f^{\mathrm{ta}}(\mathbf{X}_t+\epsilon_t+\sum_{t-L}^{t-1}\epsilon_\tau,\mathbf{T})+\lambda\|\Gamma\|_{2,1}.
	\end{align}

    We use the sign gradient descent to minimize the two objective functions, \ie, Eq.~\ref{eq:uaoia_label} and \ref{eq:taoia_label}, with the step size of 0.3, followed by a clip operation.
    In Eq.~\ref{eq:uaoia_label} and \ref{eq:taoia_label}, $\lambda$ controls the regularization degree and we set it to a constant 0.00001.
	Online minimizing Eq.~\ref{eq:uaoia_label} and \ref{eq:taoia_label} can be effective and efficient. First, optimizing the incremental perturbation is equivalent to optimizing $\mathbf{E}_t$ by regarding $\mathbf{E}_{t-1}$ as the start point. Since neighboring frames of a video is usually similar, such start point helps get an effective perturbation within very few iterations. Second, the $L_{2,1}$ norm make incremental perturbations to be spatial-temporal sparse and let $\mathbf{E}_t$ to be more imperceptible. For example, when applying SPARK on the SiamRPN-AlexNet-based trackers, we find following observations:
	
        {\bf Spatial-temporal sparsity of incremental perturbations:}  
			%with the $L_{2,1}$ regularization in Eq.~(\ref{eq:oim_func}), 
			The incremental perturbations become gradually sparse along the space and time (see Fig.~\ref{fig:analysis}~(d)). 
			This facilitates generating more imperceptible perturbations than BA methods.
			In addition, SPARK gets the smallest MAP across all frames with higher success rate than BA-E on OTB100 (see Fig.~\ref{fig:analysis}~(b)).
		
		{\bf Efficient optimization}: Fig.~\ref{fig:analysis}~(d) depicts the loss values during optimization from frame $41$ to $49$.
		At frame $41$, it takes about 7 iterations to converge. 
        However, at other frames, we obtain minimum loss in only two iterations.
		It enables more efficient attack than BA methods.
		As presented in Table~\ref{tab:analysis_results}, SPARK only uses 2.25 iterations at average to achieve 78.9\% success rate.

	The sparsity and efficiency of SPARK potentially avoid high-cost iterations at each frame. 
	In practice, we perform SPARK at every 30 frames\footnote{We use 30 as the attack interval since videos are usually at 30 fps and such setup naturally utilizes the potential delay between 29th and 30th frames.} and calculate $\mathbf{E}_{t_0}$ by optimizing Eq.~\ref{eq:uaoia_label} or Eq.~\ref{eq:taoia_label} with 10 iterations.
	In addition, we attack on the search region of the attacked tracker instead of the whole frame to accelerate the attacking speed. The search region of the $t$th frame is cropped from $\mathbf{X}_t$ at the center of predictive result of frame $t-1$, \ie, $\mathbf{b}^{\mathrm{a}}_{t-1}$, and the trackers can be reformulated as $\phi_{\theta{'}}(\mathbf{X}_t,\mathbf{T}, \mathbf{b}_{t-1}^{\mathrm{a}})$ and $\phi_{\theta}(\mathbf{X}_t,\mathbf{T}, \mathbf{b}_{t-1}^{\mathrm{a}})$. 
	We will discuss the attack results without $\mathbf{b}_{t-1}^{\mathrm{a}}$ in the experiments.
	\begin{wrapfigure}[11]{r}{0.5\textwidth}\vspace{-1em} 
	\begin{center}
		\includegraphics[width=1.0\linewidth]{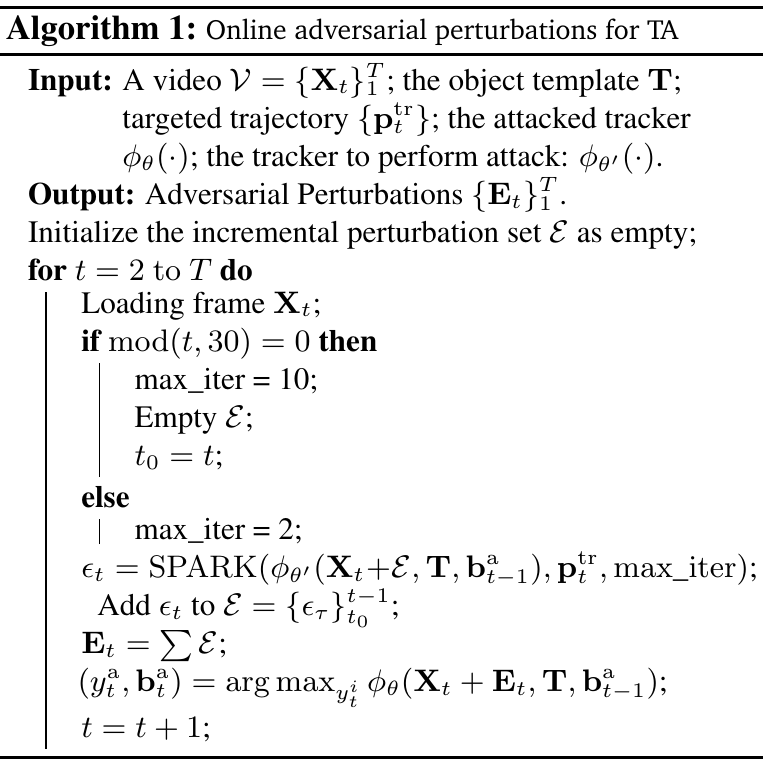}
	\end{center}
	\label{fig:algo_spark}
    \end{wrapfigure}
	We perform both UA and TA against visual tracking and summarize the attack process of SPARK for TA in Algorithm~\textcolor{red}{1}. 
	At frame $t$, we first load a clean frame $\mathbf{X}_t$. 
	If $t$ cannot be evenly divisible by 30, we optimize the objective function, \ie, Eq.~\ref{eq:taoia_label}, with 2 iterations and get $\epsilon_t$. 
	Then, we add $\epsilon_t$ into $\mathcal{E}$ that stores previous incremental perturbations, \ie, $\{\epsilon_\tau\}_{t_0}^{t-1}$, and obtain $\mathbf{E}_t=\sum\mathcal{E}$. 
	If $t$ can be evenly divisible by 30, we clear $\mathcal{E}$ and start a new round attack.

%-------------------------------------------------------------------------
%-------------------------------------------------------------------------
%\vspace{-1.5em}
\section{Experimental Results}

%-------------------------------------------------------------------------
%\vspace{-1em}
\subsection{Setting}
%\vspace{-0.5em} 
%
\textbf{Datasets.} We select 4 widely used datasets, \ie, \textbf{OTB100} \cite{Wu15}, \textbf{VOT2018} \cite{Kristan2018a}, \textbf{UAV123} \cite{Mueller2016ECCV}, and \textbf{LaSOT} \cite{Fan2019LaSOT} as subject datasets. 
OTB100 and VOT2018 are general datasets that contain 100 videos and 60 videos.  
UAV123 focuses on videos captured by UAV and includes 123 videos and LaSOT is a large scale dataset containing 280 testing videos.

\textbf{Models.} Siamese network~\cite{Bertinetto16-2,Guo17_ICCV,Li2018CVPR,Zhu2018ECCV,Li2019CVPR,Fan2019CVPR} is a dominant tracking scheme that achieves top accuracy with beyond real-time speed.  
We select SiamRPN-based trackers \cite{Li2018CVPR,Li2019CVPR} that use AlexNet \cite{Krizhevsky2012NIPS}, MobileNetv2 \cite{Howard2017arxiv}, and ResNet-50 \cite{He2016CVPR} as backbones, since they are built on the same pipeline and achieve the state-of-the-art performance on various benchmarks.
We also study the attacks on online udpating variants of SiamRPN-based trackers and the SiamDW tracker~\cite{Zhang2019CVPR}.

\textbf{Metrics.} 
We evaluate the effectiveness of adversarial perturbations on the basis of center location error~(CLE) between predicted bounding boxes and the ground truth or targeted positions.
In particular, given the bounding box annotation at frame $t$, \ie, $\mathbf{b}_t^\mathrm{an}$, we say that a tracker locates an object successfully, if we have $\mathrm{CLE}(\mathbf{b}_t, \mathbf{b}_t^\mathrm{an})=\|ce(\mathbf{b}_t)-ce(\mathbf{b}_t^\mathrm{an})\|_2<20$ where $\mathbf{b}_t$ is the predicted box~\cite{Wu15}.
Similarly, we say an attacker succeeds at frame $t$ when $\|ce(\mathbf{b}_t)-\mathbf{p}_t^\mathrm{tr}\|_2<20$ where $\mathbf{p}_t^\mathrm{tr}$ is the $t$th position on a given targeted trajectory.
With above notations, we define precision drop for UA, success rate for TA, and MAP for both UA and TA: (1) \textbf{Prec.\,Drop:} Following \cite{Wei2019TransferableAA} and \cite{Xie2017AdversarialEF}, for UA, we use precision drop of a tracker (after attacking) to evaluate the generated adversarial perturbations. The precision of a tracker is the rate of frames where the tracker can locate the object successfully. (2) \textbf{Succ.\,Rate:} For TA, Succ.\,Rate denotes the rate of frames where an attack method fools a tracker successfully. (3) \textbf{MAP:} Following \cite{Wei2018SparseAP}, we use the mean absolute perturbation~(MAP) to measure the distortion of adversarial perturbations. For a video dataset containing $D$ videos, we have $\mathrm{MAP}=\frac{1}{D*K}\sum_d\sum_k\frac{1}{M*C}\sum_i\sum_c|\mathbf{E}_{k,d}(i,c)|$, where $K$, $M$ and $C$ refer to the number of frames, pixels and channels, respectively.

%
%\subsection{Untarged and targeted attack results~(RQ1)}

%\subsubsection{Quantitative results.}
%
%\subsubsection{Visualization results.}

\textbf{Configuration.} For TA, the targeted trajectory, \ie, $\{\mathbf{p}^{\mathrm{tr}}_t\}_1^T$, is constructed by adding random offset values to the targeted position of previous frame, \ie, $\mathbf{p}^{\mathrm{tr}}_t=\mathbf{p}^{\mathrm{tr}}_{t-1}+\Delta\mathbf{p}$, where $\Delta\mathbf{p}$ is in the range of 1 to 10. 
The generated trajectories are often more challenging than manual ones due to their irregular shapes.

\begin{wrapfigure}[12]{r}{0.5\textwidth}\vspace{-3em} 
	\begin{center}
		\includegraphics[width=1.0\linewidth]{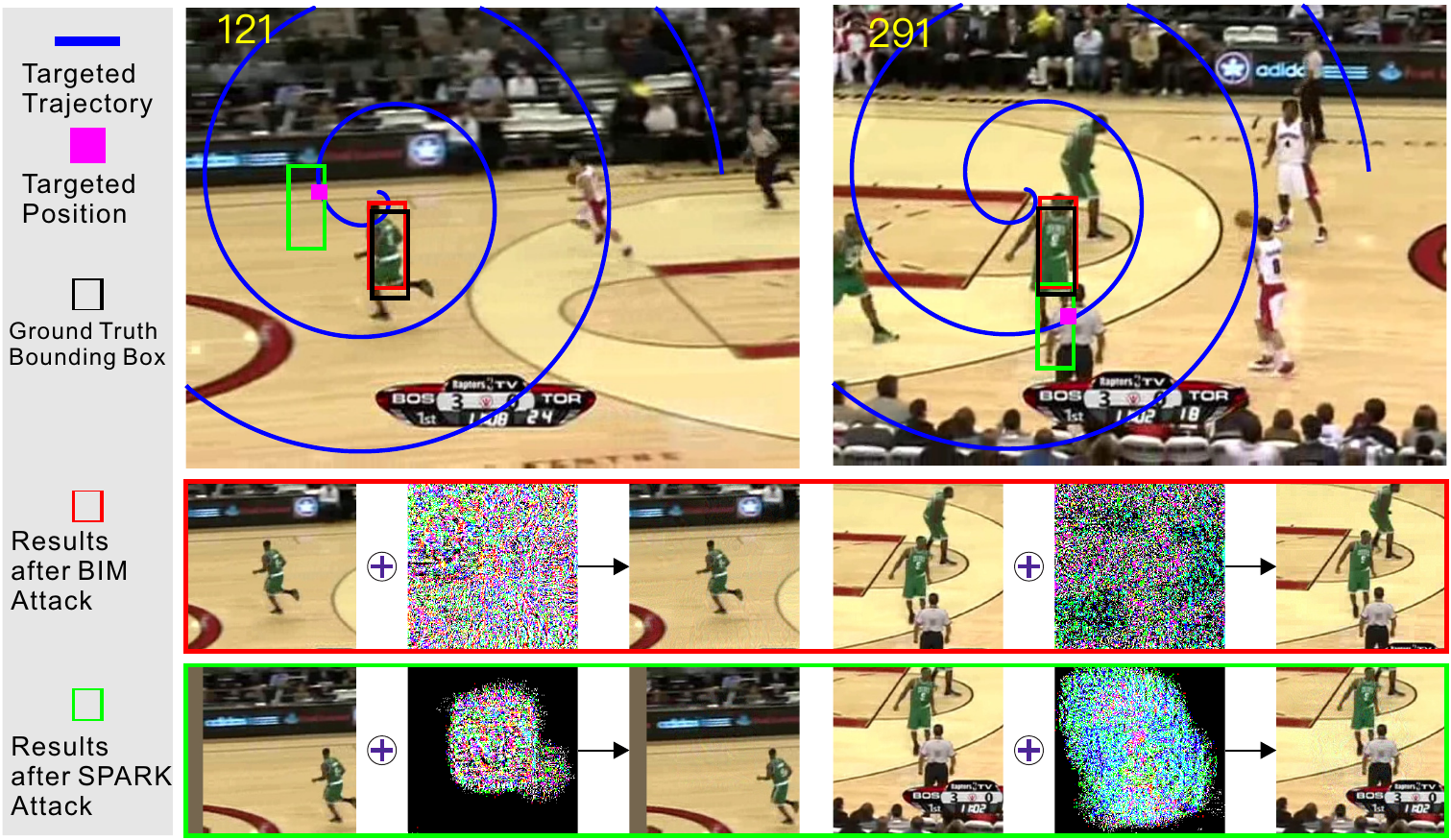}
	\end{center}\vspace{-1em} 
	\caption{\fontfamily{bch}\selectfont {An TA example based on BIM and SPARK. We use a spiral line as the targeted trajectory that embraces the object at most of the time and makes the TA challenge. }}%
	\label{fig:case}
\end{wrapfigure}

%-------------------------------------------------------------------------
%\vspace{-1.5em}
\subsection{Comparison Results}
%\vspace{-0.5em} 

\textbf{Baselines.} 
Up to present, there still lacks research about adversarial attack on online object tracking. Therefore, we compare with baselines by constructing basic attacks and extending the existing video attack technique. 
To further demonstrate the advantages of SPARK over existing methods, we extend the BA-E in Table~\ref{tab:analysis_results} such that it has the same configuration with SPARK for a more fair comparison. 
To be specific, original BA-E attacks each frame with 10 iterations. 
However, in Algorithm~\textcolor{red}{1}, SPARK attacks every 30 frames with 10 iterations while the frame in interval are attacked with only 2 iterations. We configure the new BA-E with the similar iteration strategy and adopt different optimization methods (\ie, FGSM, BIM \cite{BIM_2016_ICLRW}, MI-FGSM \cite{Dong2018BoostingAA}, and C\&W).
In addition, we tried our best to compare with the existing method, \ie, \cite{Wei2018SparseAP} designed for action recognition. 
However, it uses all frames of a video to predict the category and cannot directly be used for attacking online tracking.
We made an extension of it, \ie, when attacking at frame $t$, the previous 30 frames are used to generate the adversarial.

\begin{table*}[t]
    \fontfamily{bch}\selectfont
	\caption{\fontfamily{bch}\selectfont {Attacking three models with proposed SPARK method on OTB100 and VOT2018 for both UA and TA. The comparison results of 5 existing attack methods are also reported. The results on two larger datasets, \ie, UAV123 and LaSOT, for attacking SiamRPN-AlexNet are presented. The best three results are highlighted by \first{red}, \second{green}, and \third{blue}, respectively.}}%\vspace{-1em}
	\tiny
	\setlength{\tabcolsep}{1.5pt}
	\begin{center}
		\begin{tabular}{l|c|c|c|c|c|c|c|c|c|c|c}
			\hline
			\multirow{3}{*}{SiamRPN} & \multirow{3}{*}{Attacks}  &\multicolumn{6}{|c|}{Untargreted Attack~(UA)} & \multicolumn{4}{|c}{Targeted Attack~(TA)}\\
			\cline{3-12}
			& &\multicolumn{3}{|c|}{{OTB100}} & \multicolumn{3}{|c|}{{VOT2018}} &\multicolumn{2}{|c|}{{OTB100}} & \multicolumn{2}{|c}{{VOT2018}}  \\
			\cline{3-12}
			& & { Org. Prec.~(\%)} & {Prec. Drop~(\%)} & {MAP}  & {Org. Prec.~(\%)} & {Prec. Drop~(\%)} & {MAP}  & {Succ. Rate~(\%)} & {MAP} & {Succ. Rate~(\%)} & {MAP} \\
			\hline
			\hline
			\multirow{6}{*}{AlexNet} & FGSM & 85.3 & 8.0 & 1.24 & 65.8 & 13.6 & 1.24 & 7.9 & 1.24 & 4.3 & 1.24 \\
			\cline{2-12}
			& BIM & 85.3 & \second{72.1} & 2.17 & 65.8 & \third{57.4} & 2.28 & \third{38.8} & 2.14 & \second{48.5} & 2.10 \\
			\cline{2-12}
			& MI-FGSM & 85.3 & \third{68.4} & 3.70 & 65.8 & \second{58.2} & 4.31 & \second{41.8} & 3.18 &  \third{47.0} & 3.17 \\
			\cline{2-12}
			& C\&W & 85.3 & 54.2 & 1.31 & 65.8 & 50.6 & 1.26 & 25.7 & 1.27 & 25.7 & 1.23 \\
			\cline{2-12}
			& Wei & 85.3 & 25.9 & 0.21 & 65.8 & 33.6 & 0.30 & 16.0 & 0.27 & 20.9 & 0.24 \\
			\cline{2-12}
			& {\bf SPARK} & 85.3 & \first{78.9}  & 1.04 & 65.8 & \first{61.6}  & 1.03 & \first{74.6} & 1.36 & \first{78.9} & 1.38 \\
			\hline
			\hline
			\multirow{6}{*}{Mob.} & FGSM & 86.4 & 6.7 & 1.00 & 69.3 & 14.1 & 0.99 & 7.9 & 1.00 & 3.4 & 0.99 \\
			\cline{2-12}
			& BIM & 86.4 & 37.8 & 1.07 & 69.3 & \third{46.2} & 1.06 & \third{30.3} & 1.06 & \second{32.9} & 1.05 \\
			\cline{2-12}
			& MI-FGSM & 86.4 & \second{42.3} & 1.71 & 69.3 & \second{46.6} & 1.73 & \second{33.5} &  1.70 &  \third{32.7} & 1.71 \\
			\cline{2-12}
			& C\&W & 86.4 & 23.6 & 1.04 & 69.3 & 28.2 & 1.02 & 13.7 & 1.05 & 8.9 & 1.01 \\
			\cline{2-12}
			& Wei & 86.4 & \third{39.4} & 0.84 & 69.3 & 27.8 & 0.54 & 11.3 & 0.51 & 7.0 & 0.53  \\
			\cline{2-12}
			& {\bf SPARK} & 86.4 &  \first{54.1} & 1.66 & 69.3 & \first{55.5} & 1.25  & \first{51.4} & 1.65 & \first{45.5} & 1.21 \\
			\hline
			\hline
			\multirow{6}{*}{Res50} & FGSM & 87.8 & 4.5 & 0.99 & 72.8 & 8.1 & 0.99 & 7.7 & 0.92 & 2.9 & 0.99 \\
			\cline{2-12}
			& BIM & 87.8 & \third{27.0} & 1.10 & 72.8 & \third{39.1} & 1.10 & \third{17.1} & 1.09 & \third{17.0} & 1.08 \\
			\cline{2-12}
			& MI-FGSM & 87.8 & \first{31.9} & 1.72 & 72.8 & \second{41.8} & 1.75 & \second{18.8} &  1.71 &  \second{19.5} & 1.72 \\
			\cline{2-12}
			& C\&W & 87.8 & 14.6 & 1.03 & 72.8 & 20.4 & 1.01 & 10.0 & 1.04 & 5.3 & 1.01 \\
			\cline{2-12}
			& Wei & 87.8 & 9.7 & 0.65 & 72.8 & 15.7 & 0.68 & 9.7 & 0.78 & 4.8 & 0.69 \\
			\cline{2-12}
			& {\bf SPARK} &87.8 & \second{29.8} & 1.67 & 72.8  & \first{54.3} & 1.26 & \first{23.8} & 1.70 & \first{39.5} & 1.26\\
			\hline
			\hline
			\multirow{3}{*}{SiamRPN} & \multirow{3}{*}{Attacks}  &\multicolumn{6}{|c|}{Untargreted Attack~(UA)} & \multicolumn{4}{|c}{Targreted Attack~(TA)}\\
			\cline{3-12}
			& &\multicolumn{3}{|c|}{{UAV123}} & \multicolumn{3}{|c|}{{LaSOT}} &\multicolumn{2}{|c|}{{UAV123}} & \multicolumn{2}{|c}{{LaSOT}}  \\
			\cline{3-12}
			& & { Org. Prec.} & {Prec. Drop} & {MAP}  & {Org. Prec.} & {Prec. Drop} & {MAP}  & {Succ. Rate} & {MAP} & {Succ. Rate} & {MAP} \\
			\hline
			\hline
			\multirow{6}{*}{AlexNet} & FGSM & 76.9 & 3.7 & 1.25 & 43.5 & 4.0 & 1.22 & 3.7 & 1.25 & 4.70 & 1.22 \\
			\cline{2-12}
			& BIM & 76.9 & \second{36.4} & 1.70 & 43.5 & \second{32.0} & 1.64 & \second{28.7} & 1.75 & \third{17.4} & 1.73 \\
			\cline{2-12}
			& MI-FGSM & 76.9 & \third{31.5} & 2.54 & 43.5 & \third{31.6} & 2.50 & \third{28.3} & 2.53 &  \second{17.8} & 2.46 \\
			\cline{2-12}
			& C\&W & 76.9 & 17.0 & 1.37 & 43.5 & 19.9 & 1.29 & 11.0 & 1.36 & 8.7 & 1.28 \\
			\cline{2-12}
			& Wei & 76.9 & 5.6 & 0.31 & 43.5 & 9.3 & 0.29 & 6.8 & 0.37 & 6.9 & 0.31 \\
			\cline{2-12}
			& {\bf SPARK} & 76.9 & \first{43.6}  & 1.13 & 43.5 & \first{38.2}  & 0.93 & \first{54.8} & 1.06 & \first{48.9} & 1.09 \\
			\hline
		\end{tabular}
	\end{center}
	\label{tab:com_results}%\vspace{-6em}
\end{table*}

\textbf{Results.} Table~\ref{tab:com_results} shows the TA/UA results on the four datasets. Column \textit{Org.~Prec.} gives the precision of the original tracker. Due to the large evaluation effort, for UAV123 and LaSOT, we only perform the more comprehensive comparison on the smaller model, \ie, SiamRPN-AlexNet.

We observe that: 
1) Compared with the existing attacks, SPARK achieves the highest Prec.~Drop for UA and Succ. Rate for TA on most of datasets and models. For the results of attacking SiamRPN-Res50 on OTB100, SPARK gets slightly smaller Proc.~Drop than MI-FGSM but generates more imperceptible perturbations. 
2) SPARK generates imperceptible perturbations. When attacking SiamRPN-AlexNet on all datasets, SPARK always gets more imperceptible perturbations than FGSM, BIM, MI-FGSM, and C\&W. \cite{Wei2018SparseAP} produces the smallest perturbations but the attacking is not effective. 
%For example, it only achieves 25.9\% precision drop in UA and 16.0\% success rate in TA on the OTB100 dataset. 
Similar results can be also found on other three datasets. 
%attacking deeper models, the perturbations of SPARK become less imperceptible. Nevertheless, compared with MIFGSM SPARK always generates more imperceptible perturbations with larger precision drop or success rate.
%
3) In general, it is more difficult to attack deeper models for all attacks, since the Prec.~Drop and Succ.~Rate of almost all attacks gradually become smaller as the models become more complex.

In summary, the results of Table~\ref{tab:analysis_results} and \ref{tab:com_results} indicate the effectiveness of SPARK in attacking the tracking models with small distortions. In addition to the quantitative results, we give a concrete example base on BIM and SPARK~(see Fig.~\ref{fig:case}). Compared with BIM, SPARK lets the SiamRPN-AlexNet tracker always produces bounding boxes on the targeted trajectory with a sparse perturbation, indicating the effectiveness of SPARK. 
\subsection{Analysis of SPARK}
%\vspace{-0.5em} 
\label{subsec:exp_analysis}
\textbf{Validation of the online incremental attack.} We implement six variants of SPARK by setting $L\in\{5,10,15,20,25,30\}$ in Eq.~\ref{eq:taoia_label} to analyze how historical incremental perturbations affect attacking results. For example, when attacking the frame $t$ with $L=5$, we use previous 5 incremental perturbations to generate $\mathbf{E}_t$. We use these SPARKs to attack SiamRPN-AlexNet under TA on OTB100 and report the Succ. Rate, MAP, and MAP difference~(MAP Diff.($L$)) in Fig.~\ref{fig:analysis}, where MAP Diff.($L$)=MAP(SPARK($L$))-MAP(SPARK($L-1$)). We see that: 1) the Succ. Rate increases with the growing of $L$. It demonstrates that historical incremental perturbations do help achieve more effective attack. 2) Although MAP also gets larger as the $L$ increases, the MAP Diff. gradually decrease. This validates the advantages of SPARK, that is, it can not only leverage temporal transferability effectively but also maintaining the imperceptible perturbations. 

\begin{figure}[h]
	\begin{center}
		\includegraphics[width=0.98\linewidth]{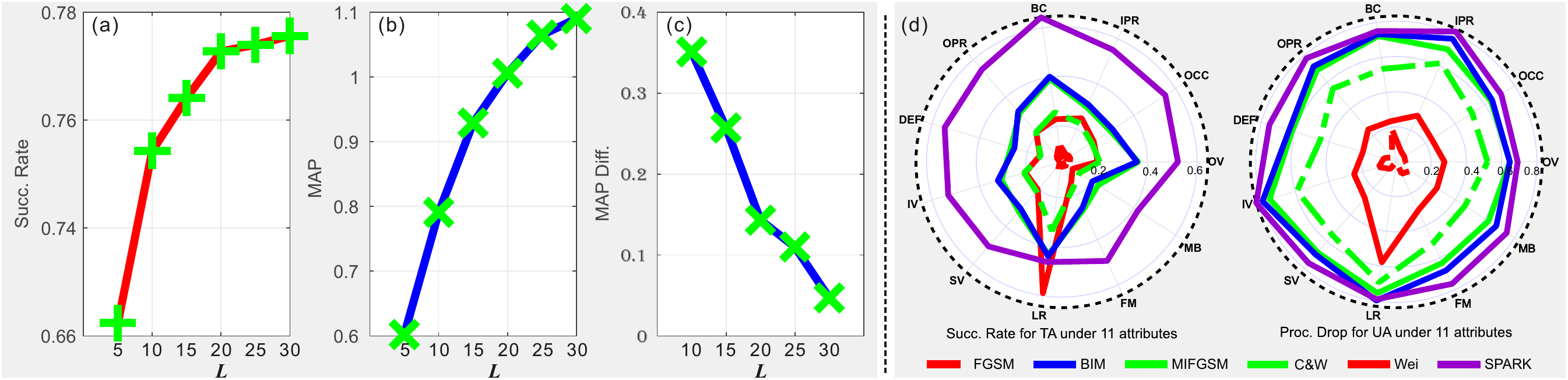}%\vspace{-2em}
	\end{center}
	\caption{\fontfamily{bch}\selectfont {(a) and (b) are the Succ. Rate and MAP of six variants of SPARK under TA for SiamRPN-AlexNet. The six variants are built by using different number of previous perturbations for Eq.~(\ref{eq:taoia_label}) and (c) shows the MAP difference between neighboring variants.(d) Attacking SiamRPN-AlexNet with the six compared methods on the 11 subsets of OTB100 for both TA and UA.}}
	\label{fig:attr}%\vspace{-2.5em}
\end{figure}

\textbf{Results under Challenging Attributes.} OTB dataset contains 11 subsets corresponding to 11 interference attributes\footnote{\tiny{The 11 attributes are illumination variation~(IV), scale variation~(SV), in-plane rotation~(IPR), outplane rotation~(OPR), deformation~(DEF), occlusion~(OCC), motion blur~(MB), fast motion~(FM), background clutter~(BC), out-of-view~(OV), and low resolution~(LR).}}. Fig.~\ref{fig:attr} shows results of six methods for SiamRPN-AlexNet on 11 subsets. We observe that: 1) SPARK has much larger Prec. Drop and Succ. Rate than baselines on all subsets except the LR one for both UA and TA. 2) The advantages of SPARK over baselines for TA is more significant than that for UA. 3) BIM, Wei, MIFGSM, and C\&W are much more effective under the LR attribute than others. This may be caused by the limited effective information in LR frames, which leads to less discriminative deep representation and lets the attacking more easier.

\begin{table}[t]
    \fontfamily{bch}\selectfont
	\caption{\fontfamily{bch}\selectfont {Left sub-table shows the results of attacking DSiamRPN trackers on OTB100 for UA and TA while the right one presents the results of attacking SiamDW trackers.}}%\vspace{-1em}
	\tiny
	\setlength{\tabcolsep}{1.5pt}
	\begin{center}
		\begin{tabular}{l|c|c|c|c|c|c|c|c}
			\hline
			\multirow{2}{*}{} & \multicolumn{2}{c|}{UA Attack} & TA Attack & & \multirow{2}{*}{} & \multicolumn{2}{c|}{UA Attack} & TA Attack \\
			\cline{2-4}\cline{7-9}
			& Org. Prec.(\%) & Prec. Drop(\%) & Succ. Rate(\%) & & & Org. Prec.(\%) & Prec. Drop(\%) & Succ. Rate(\%) \\
			\hline
			\hline
			DSiam-AlexNet & 86.6 & 78.5 & 65.9 & & SiamDW-CIResNet & 83.0 & 58.1 & 21.5 \\
			\cline{1-4}\cline{6-9}
			DSiam-Mob. & 87.8 & 56.8 & 44.4 & & SiamDW-CIResNext & 81.7 & 74.2 &29.4 \\
			\cline{1-4}\cline{6-9}
			DSiam-Res50 & 90.3  & 37.1 & 20.4 & & SiamDW-CIResIncep & 82.3 & 70.2 & 30.8 \\
			\hline
		\end{tabular}
	\end{center}
	\label{tab:online_results}%\vspace{-4em}
\end{table}

\textbf{Transferability across Models.} We discuss the transferability across models, which is to apply perturbations generated from one model to another. 
In Table~\ref{tab:across_model}, the values in the UA and TA parts are the \emph{Prec. Drop} and \emph{Succ. Rate}, respectively.
We see that the transferability across models also exists in attacking object tracking. All attack methods lead to the precision drop to some extent. 
For example, the perturbations generated by SiamRPN-Res50 cause the precision of SiamRPN-Mob. drop 16.1, which is a huge performance degradation in tracking evaluation.
For TA, after transferability, the success rate is around 6.5 for all cases. 
Such limited transferability may be caused by the insufficient iterations during online process and can be further studied in the future.

\textbf{SPARK without object template $\mathbf{T}$.} As discussed in Section~\ref{subsec:oia} and Algorithm~\textcolor{red}{1}, the tracked object, \ie, the template $\mathbf{T}$, should be given during attack. Here, we demonstrate that we can realize effective attack without $\mathbf{T}$. Specifically, given the first frame of an online video, we use SSD~\cite{Liu2016} to detect all possible objects in the frame and select the object nearest to the frame center as the target object. The basic principle behind this is that a tracker usually starts working when the object is within the camera's center view. As presented in Table~\ref{tab:across_model}, without the specified $\mathbf{T}$, SPARK-no$\mathbf{T}$ also acheive 71.0\% Prec.~Drop under UA on OTB100 and is slightly lower than the original SPARK.

\textbf{SPARK without the attacked tracker's predictions.} 
In Algorithm~\textcolor{red}{1}, we detail that our SPARK is performed on the search region of $\phi_{\theta}(\cdot)$ and require the attacked tracker's prediction, \ie, $\mathbf{b}^\mathrm{a}_{t-1}$, as an additional input, which might limit the application of our method since we might not access to the attacked tracker's predictions. A simple solution is to replace the $\mathbf{b}^\mathrm{a}_{t-1}$ in the Algorithm~\textcolor{red}{1} with $\mathbf{b}^\mathrm{a{'}}_{t-1}$, \ie, we can perform attack on the search region of $\phi_{\theta{'}}(\cdot)$ and propagate the perturbations to the whole frame. As shown in Table~\ref{tab:across_model}, without the attacked tracker's predictions, SPARK-no$\mathbf{b}^{\mathrm{a}}_t$ gets 67.7\% Prec.~Drops under UA on OTB100 which is slightly lower than the original SPARK.
			
%-------------------------------------------------------------------------
\begin{table}[t]
    \fontfamily{bch}\selectfont
	\caption{\fontfamily{bch}\selectfont { The left subtable shows the transferability between subject models~(\ie, AlexNet, MobileNetv2, and ResNet50) on OTB100. Values in UA and TA are Proc. Drop and Succ. Rate, respectively. The right subtable shows the results of attacking SiamRPN-AlexNet on OTB100 without object template $\mathbf{T}$ or 
	attacked tracker's prediction $\mathbf{b}^{\mathrm{a}}_t$.
	The third row of this subtable is the original results of SPARK in Table~\ref{tab:com_results}. }}%\vspace{-1em}
	\tiny
	\setlength{\tabcolsep}{1.5pt}
	\begin{center}
		\begin{tabular}{l|c|c|c|c|c|c|c|c|l|c|c|c}
			\hline
			\multirow{2}{*}{} & \multicolumn{3}{c|}{Proc.~Drop of UA from } & & \multicolumn{3}{c|}{Succ.~Rate of TA from} & &\multirow{2}{*}{}  &\multicolumn{2}{c|}{Untargreted Attack~(UA)} & Targeted Attack~(TA) \\
			\cline{2-4}\cline{6-8}\cline{11-13}
			& AlexNet & Mob.Net & Res50 & & AlexNet & Mob.Net & Res50 & & & Org.~Prec. & Prec.~Drop & Succ. Rate \\
			\hline
			\hline
			SiamRPN-AlexNet & 78.9 & 6.7 & 2.0 & & 74.6 & 6.2 & 6.7 & & SPARK-no$\mathbf{T}$ & 85.3 & 71.0 & 50.6 \\
			\cline{1-4}\cline{6-8}\cline{10-13}
			SiamRPN-Mob. & 3.5 & 54.1 & 2.7 & & 6.3 & 51.4 & 6.6 & & SPARK-no$\mathbf{b}^{\mathrm{a}}_t$ & 85.3 & 67.7 & 46.2\\
			\cline{1-4}\cline{6-8}\cline{10-13}
			SiamRPN-Res50 & 7.5 & 16.1 & 29.8 & & 6.2 & 6.5 & 23.8 & & SPARK & 85.3 & 78.9 & 74.6 \\
			\hline
		\end{tabular}
	\end{center}
	\label{tab:across_model}%\vspace{-5.5em}
\end{table}

%-------------------------------------------------------------------------
%\vspace{-1.5em}
\subsection{Attacking other tracking frameworks}
%\vspace{-0.5em} %~(RQ4)}
%
%In tracking area, it has been known that online updating module, e.g., correlation filter~(CF), helps trackers to adapt target and background changes.
%
%It is interesting to explore how adversarial perturbations would affect the correlation filter~(CF)-based trackers. 
%
%To this end, 
\textbf{Transferability to online updating trackers.}
We construct three online updating trackers with dynamic Siamese tracking~(DSiam)~\cite{Guo17_ICCV}, and obtain trackers: DSiamRPN-AlexNet, MobileNetV2, and ResNet-50. 
We then use the adversarial perturbations from SiamRPN-AlexNet, MobileNetV2, and ResNet-50 to attack the DSiamRPN-based trackers.
In Table~\ref{tab:online_results}, we observe that: 1) DSiam indeed improves the precision of three SiamRPN-based trackers according to the results in Table~\ref{tab:com_results}.
2) The adversarial perturbations from SiamRPNs is still effective for DSaim versions with the precision drops being 78.5\%, 56.8\%, and 37.1\% which are larger than the results in Table~\ref{tab:com_results}. This is because DSiamRPN-based trackers use online tracking results that may have been fooled by attacks to update models and make them less effective, thus are easily attacked.

\textbf{Attacking SiamDW~\cite{Zhang2019CVPR}.} We validate the generality of SPARK by attacking another tracker, \ie, SiamDW~\cite{Zhang2019CVPR} that won the VOT-19 RGB-D challenge and achieved the runner-ups in VOT-19 Long-term and RGB-T challenges~\cite{Kristan2019a}. As shown in right sub-table of the Table~\ref{tab:online_results}, without changing any attack parameters, SPARK significantly reduces the precision of SiamDW trackers under the UA, demonstrating its generality.

%-------------------------------------------------------------------------
%-------------------------------------------------------------------------
%\vspace{-1.5em}
\section{Conclusion}
%\vspace{-1em}

In this paper, we explored adversarial perturbations for misleading the online visual object tracking along an incorrect~(untarged attack, UA) or specified~(targeted attack, TA) trajectory. 
An optimization-based method, namely \textit{\underline{sp}atial-\underline{a}ware online inc\underline{r}emental attac\underline{k}} (SPARK), was proposed to overcome the challenges introduced in this new task.
SPARK optimizes perturbations with a $L_{2,1}$ regularization norm and considers the influence of historical attacking results, thus is more effective.
% significantly improved the imperceptibility and can be efficiently optimized in average 2.25 iterations perframe. 
%
Experimental results on OTB100, VOT2018, UAV123, and LaSOT showed that SPARK successfully fool the state-of-the-art trackers.

%\vspace{-1.5em}
\section{Acknowledgements}
%\vspace{-1em}
This work was supported by the National Natural Science Foundation of China (NSFC) under Grant 61671325, Grant 61572354, Grant 61672376, Grant U1803264, and Grant 61906135, the Singapore National Research Foundation under the National Cybersecurity R\&D Program No. NRF2018NCR-NCR005-0001 and the NRF Investigatorship No. NRFI06-2020-0022, and the
 National Satellite of Excellence in Trustworthy Software System No. NRF2018NCR-NSOE003-0001.
We also gratefully acknowledge the support of NVIDIA AI Tech Center (NVAITC) to our research.

\section{Supplementary Material}

%-------------------------------------------------------------------------
%-------------------------------------------------------------------------
\subsection{Attacking Correlation Filter-based Trackers}
Correlation filter~(CF) is a dominant tracking framework that can achieves well balance between tracking speed and accuracy. However, most of the CF-based trackers are not end-to-end architectures and use hand-craft features. Hence, it is difficult to attack them via the white-box setup and is meaningful to explore if SPARK could attack CF-based trackers by using deep tracking frameworks, \eg, SiamRPN-based trackers. As shown in Table~\ref{tab:cfattack}, the adversarial examples from SiamRPN-Alex can reduce all tested CF-based trackers having different features, which demonstrates that the transiferability of our attack across different trackers and features exists. In terms of different features, the HOG feature is easier attacked when compared with the gray feature, hybird feature~(\ie, HOG+CN), and deep feature~(\eg, VGG).
\vspace{-1em}
\begin{table}[h]
    \renewcommand\thetable{I}
	\caption{\small{Untargeted attack~(UA) for correlation filter-based trackers, \eg, MOSSE~\cite{Bolme10}, KCF~\cite{Henriques15}, BACF~\cite{Galoogahi17}, STRCF~\cite{Li2018CVPR}, and ECO~\cite{Danelljan16ECO} with the perturbations generated from SiamRPN-AlexNet.}}\vspace{-1em}
	\small
	\begin{center}
		\begin{tabular}{l|c|c|c|c|c}
			\hline
			& MOSSE & KCF & BACF & STRCF & ECO \\
			\hline
			\hline
			Features  & Gray & HOG & HOG+CN & HOG+CN & VGG \\
			\hline
			Org.~Prec.~(\%) & 41.7  & 69.2 & 70.5 & 72.3 & 89.6\\
			\hline
			Proc.~Drop~(\%) & 0.2 & 3.3 & 2.1 & 1.5 & 0.9 \\
			\hline
		\end{tabular}
	\end{center}
	\label{tab:cfattack}\vspace{-4em}
\end{table}
%
\if 0
\vspace{-2em}
\begin{table}[h]
	\caption{\small{Untargeted attack~(UA) for correlation filter-based trackers, \eg, MOSSE~\cite{Bolme10}, KCF~\cite{Henriques15}, BACF~\cite{Galoogahi17}, and STRCF~\cite{Li2018CVPR}, with the perturbations generated from SiamRPN-AlexNet, MobileNetv2, and Res50.}}\vspace{-1em}
	\small
	\begin{center}
		\begin{tabular}{l|c|c|c|c|c}
			\hline
			\multirow{2}{*}{} & \multirow{2}{*}{Features} & \multirow{2}{*}{Org.~Prec} & \multicolumn{3}{c}{Proc.~Drop of UA from SiamRPN-}   \\
			\cline{4-6}
			&  &  & AlexNet & MobileNet & Res50  \\
			\hline
			\hline
			MOSSE  & Gray & 41.7 & 41.5 & 37.6 & 42.1 \\
			\hline
			KCF & HOG & 69.2 & 65.9 & 63.9 & 64.6 \\
			\hline
			BACF & HOG+CN & 70.5 & 68.4 & 68.4 & 16.8 \\
			\hline
			STRCF  & HOG+CN & 72.3 & 70.8 & 8.0 & 6.4 \\
			\hline
		\end{tabular}
	\end{center}
	\label{tab:time_analysis}\vspace{-4em}
\end{table}
\fi
%
%-------------------------------------------------------------------------
%-------------------------------------------------------------------------
\subsection{Speed Analysis}
We have reported the time cost of our SPARK in Table.~1 in the submission and shown that SPARK is more suitable for attacking online trackers than three basic attack methods due to the balance between time cost and attack Succ.~Rate. Please find details in Section~3.3.
Compared with trackers' cost shown in Table.~\ref{tab:time_analysis}, the time cost of our attack method increases as the tracking model becomes larger under the white-box attack. In particular, when attacking SiamRPN-Alex, SPARK achieves near real-time attacking. 
Although the attack speed decreases with more complex models, the corresponding tracking speed is also slower and lets the influence of decreased attacking be smaller.
\begin{table}[h]
    \renewcommand\thetable{II}
	\caption{\small{Time cost of attacks w.r.t. different trackers on OTB100 dataset.}}\vspace{-1em}
	\small
	\begin{center}
		\begin{tabular}{l|c|c|c}
			\hline
			SiamRPN & AlexNet & MobileNetV2 & Res50  \\
			\hline
			\hline
			Track cost per frame~(ms)  & 9.3 & 37.6 & 42.1 \\
			\hline
			Attack cost per frame~(ms) & 41.4 & 126.9 & 156.3 \\
			\hline
			Track speed~(fps)  & 108.4 & 15.3 & 16.8 \\
			\hline
			Attack speed~(fps)  & 24.3 & 8.0 & 6.4 \\
			\hline
		\end{tabular}
	\end{center}
	\label{tab:time_analysis}\vspace{-2em}
\end{table}
We can reduce the high time cost of attacking larger models~(\eg, MobileNetv2 and Res50) by using the light one~(\eg, AlexNet) due to the existence of the transferability between models as discussed in Section~4.3 and Table~4. Specifically, we attack three trackers, \ie, SiamRPN-Alex/Mob./Res50, via SPARK with the adversarial perturbations generated from SiamRPN-Alex. Then, we calculate the attack's online speed as well as the three trackers' speed. As shown in the following Table.~\ref{tab:time_analysis}, the speed of SPARK base on SiamRPN-Alex can reach near real-time speed~(around 25~fps) for different trackers, which means our method is suitable for attacking real-time online trackers.
\begin{table}[h]
    \renewcommand\thetable{III}
	\caption{\small{Time cost of attacking trackers on OTB100. The adversarial perturbations are generated from SiamRPN-Alex.}}\vspace{-1em}
	\small
	\begin{center}
		\begin{tabular}{l|c|c|c}
			\hline
			SiamRPN & AlexNet & MobileNetV2 & Res50  \\
			\hline
			\hline
			Track speed~(fps)  & 108.4 & 15.3 & 16.8 \\
			\hline
			Attack speed~(fps)  & 24.3 & 23.1 & 22.7 \\
			\hline
		\end{tabular}
	\end{center}
	\label{tab:time_analysis}\vspace{-4em}
\end{table}
%

% \clearpage
% ---- Bibliography ----
%
% BibTeX users should specify bibliography style 'splncs04'.
% References will then be sorted and formatted in the correct style.
%
\bibliographystyle{splncs04}
\bibliography{AdvTrack}

\end{document}